\def\year{2018}\relax
\documentclass[letterpaper]{article} 
\usepackage{aaai18}  
\usepackage{times}  
\usepackage{helvet}  
\usepackage{courier}  
\usepackage{url}  
\usepackage{graphicx}  
\usepackage{amsmath}
\begin{document}
%
\title{Generative Adversarial Networks and Probabilistic Graph Models for Hyperspectral Image Classification}
 
\author{
Zilong Zhong and Jonathan Li\\
Mobile Sensing Lab, University of Waterloo\\
\{z26zhong, junli\}@uwaterloo.ca
}

\maketitle
\begin{abstract}
High spectral dimensionality and the shortage of annotations make hyperspectral image (HSI) classification a challenging problem. Recent studies suggest that convolutional neural networks can learn discriminative spatial features, which play a paramount role in HSI interpretation. However, most of these methods ignore the distinctive spectral-spatial characteristic of hyperspectral data. In addition, a large amount of unlabeled data remains an unexploited gold mine for efficient data use. Therefore, we proposed an integration of generative adversarial networks (GANs) and probabilistic graphical models for HSI classification. Specifically, we used a spectral-spatial generator and a discriminator to identify land cover categories of hyperspectral cubes. Moreover, to take advantage of a large amount of unlabeled data, we adopted a conditional random field to refine the preliminary classification results generated by GANs. Experimental results obtained using two commonly studied datasets demonstrate that the proposed framework achieved encouraging classification accuracy using a small number of data for training.
\end{abstract}

\section{Introduction}
\noindent Discriminative deep learning models have been used for hyperspectral image (HSI) classification and achieved very high classification accuracy in commonly studied cases. Convolutional neural networks (CNNs) have been used for hyperspectral image feature extraction and outperform traditional machine learning methods. However, both these approaches ignored the inherent difference (hundreds of spectral bands in HSIs) in spectral dimensionality between hyperspectral images and common images used in computer vision tasks (e.g. ILSVRC). Recently, a Spectral-Spatial Residual Network (SSRN) that considers the spectral-spatial characteristic of hyperspectral images by consecutively extracting spectral and spatial features and obtained state-of-art classification results \cite{Zhong}.  

Recently, generative adversarial networks (GANs) have been widely used for generating images that share the same distribution as real data, and can be used for unsupervised and semi-supervised learning \cite{Goodfellow}. Different from conventional generative methods, generative adversarial networks are not constrained by Markov fields or explicit approximation inference. For instance, a deep convolutional GAN, which consists of a discriminator and a generator, has been proposed to find a Nash equilibrium between the discriminator and the generator through model optimization \cite{Radford}. Although GANs were designed for unsupervised learning, they can be applied for semi-supervised learning  framework and delivered promising recognition results \cite{Odena}.

Probabilistic graphical models have been widely used for post-processing image segmentations and exploit the large amount of unlabeled data to enhance the classification performance \cite{Chen}. For instance, conditional random fields (CRF) that adopt a high-order term can consider more complex relationships between different spectral bands. The adoption of probabilistic graphical models is an effective way to take the unlabeled samples into account for HSI classification, because this method requires only an initialized pixel-wised prediction and raw unlabeled HSI data. 

In this work, to efficiently utilize the spectral-spatial features and abundant unlabeled data of HSIs, we tested different GANs and the combined them with conditional random fields that follow using 10 training samples for each land cover classes in two HSI datasets. 

\section{Methodology}
As shown in Fig. 1, the whole framework is composed of a discriminator, a generator, and a post-processing CRF. Inspired by the semi-supervised GAN, we proposed a spectral-spatial convolutional layer to construct both discriminators and generators. The discriminator identifies the categories and whether the input hyperspectral cubes are real or false. Additionally, the generator tries to deceive the discriminator through implicitly learning representative features and generate synthesized images. Then,
a conditional random field refines the posterior probability given raw HSIs and the preceding HSI classification results.
If we suppose the $\boldsymbol{X}$ and $\boldsymbol{Z}$ are feature tensors in discriminators and generators, respectively. Then we can formulate the two phases as follows.
\begin{equation}
\boldsymbol{X^{i+1}} = LReLU(\boldsymbol{X}^{i} * \boldsymbol{W^{i+1}} + b^{i+1}),
\end{equation}
\begin{equation}
\boldsymbol{Z^{j+1}} = ReLU(\boldsymbol{Z}^{j} *^T \boldsymbol{W^{j+1}} + b^{j+1}),
\end{equation}
where $LReLU(\cdot)$ and $ReLU(\cdot)$ denote leaky and normal rectified linear unit activation functions, respectively. $\boldsymbol{W}$
and $b$ represents weights and biases for each layers in GANs. $*$ and $*^T$ are convolutional and transposed convolutional operations.

Gradient descent based loss optimization methods perform well for supervised deep learning models. However, they are not suitable for GANS. Model optimization has been proven effective for the convergence of training GANs \cite{Salimans}.
Futhurmore, we can include labeling information to generalize the GANs for semi-supervised tasks. If we suppose $\boldsymbol{Z}=\{\boldsymbol{z}_{i}\}$ and y represent raw hyperspectral data and their corresponding labels, and $K$ is the class number. Then the model optimization based loss function are listed as follows:
\begin{equation}
L = L_{1}+L_{2}+L_{3},
\end{equation}
\begin{equation}
L_{1}=-E_{(\boldsymbol{z},y)\sim p_{d}(\boldsymbol{z},y)}\log{P_{model}(y|\boldsymbol{z},y<K+1)},
\end{equation}
\begin{equation}
L_{2}=-E_{(\boldsymbol{z},y)\sim p_{d}(\boldsymbol{z})}\log[{1-P_{model}(y=K+1|\boldsymbol{z})]},
\end{equation}
\begin{equation}
L_{3}=-E_{z\sim p_{g}}\log{P_{model}(y=K+1|\boldsymbol{z})},
\end{equation}
where $L$ is the total loss, $L_{1}$, $L_{2}$, and $L_{3}$ represent supervised loss, unsupervised loss for cheating discriminator, and unsupervised loss for generator, respectively.
We adopted a conditional random field, a graphical model that stresses the a priori continuity assumption that neighbouring HSI pixels of similar spectral signatures tend to have the same labels, to further refine the classification maps generated from the first stage. In this work, we applied a fully-connected CRF that takes all pixels in a HSI and its corresponding predicted labels in a classification map as vertices and links them together to build a graph. We calculate the graph energy $E(\cdot)$ as follows.
\begin{equation}
E(\boldsymbol{z}) = \sum_{i} U(\boldsymbol{z_{i}}) + \sum_{i,j} P(\boldsymbol{z_{i},z_{j}})),
\end{equation}
\begin{equation}
P(\boldsymbol{z_{i},z_{j}}) = \mu_{i,j}[w_1f_1(p_i,p_j,I_i,I_j)+w_2f_2(p_i,p_j)],
\end{equation}
where $U(\cdot)$ and $P(\cdot)$ are the unary and pairwise terms, respectively. $\mu$, $w_1$, $w_2$ are adjustable weights. 
$f_1(\cdot)$ and $f_2(\cdot)$ are pairwise cost that considers the spatial and spectral correlation between hyperspectral cubes. 

\begin{figure}[!t]
\centering
\includegraphics[width=3.3in]{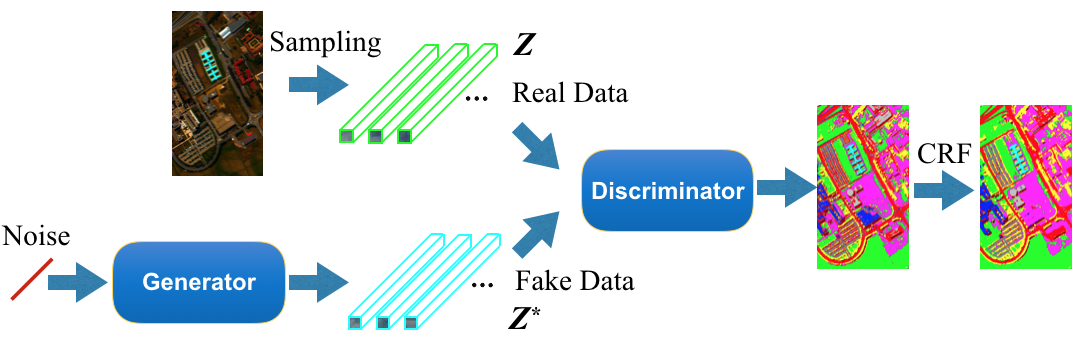}
\caption{Generative adversarial network and conditional random field based framework for HSI classification.}
\label{fig_1}
\end{figure}

\begin{table}[!t]
\renewcommand{\arraystretch}{1.3}
\caption{OA of different GANs for the IN data set}
\label{table_1}
\centering
\begin{tabular}{c | c c c c}
\hline
OA & CONV & SPA & SPC & SS \\
\hline
With CRF&$46.47\%$&$35.3\%$&$68.66\%$&$\boldsymbol{84.18\%}$\\
    Without CRF&$45.09\%$&$36.46\%$&$63.74\%$&$\boldsymbol{80.88\%}$\\
\hline
\end{tabular}
\end{table}

\begin{table}[!t]
\renewcommand{\arraystretch}{1.3}
\caption{OA of different GANs for the UP data set}
\label{table_2}
\centering
\begin{tabular}{c | c c c c}
\hline
OA & CONV & SPA & SPC & SS \\
\hline
With CRF&$66.22\%$&$39.10\%$&$73.39\%$&$\boldsymbol{83.66\%}$\\
Without CRF&$58.08\%$&$45.25\%$&$70.70\%$&$\boldsymbol{80.83\%}$\\
\hline
\end{tabular}
\end{table}

\section{Experimental Results}
To conduct our preliminary experiments, we used two commonly studied hyperspectral datasets: the Indian Pines (IN) dataset and the University of Pavia (UP) data. We adopted overall accuracy (OA) to quantitatively measure the HSI classification performance of different models.

In order to demonstrate its effectiveness, we compared the spetral-spatial (SS) GAN with other types of convolutional networks, such as autoencoder (AE) , CNN plus PCA (CNN-PCA), traditional CNN, spatial only network (SPA), and spectral only network (SPC) \cite{Zhong}. Additionally, we tested the post-processing conditional random fields on top of GANs. As demonstrated in Table 1 and 2, the proposed spectral-spatial GAN outperformed its competitors in two cases because the combination of the SS layers extracts discriminative features. However, the spatial convolutional version GAN performed not well. Furthermore, CRFs improved HSI classification accuracy in all models but the one with spatial constitutional layers.

\section{Conclusion}
Compared to traditional machine learning methods, the integration of deep learning models and
conditional random fields has achieved promising results and opened a new window for semisupervisedlearning of hyperspectral image classification. 

\bibliography{bibfile}
\bibliographystyle{aaai}

\end{document}